\documentclass[final]{cvpr}

\usepackage{times}
\usepackage{epsfig}
\usepackage{graphicx}
\usepackage{amsmath}
\usepackage{amssymb}
\usepackage[misc]{ifsym}

\usepackage{booktabs} % for pretty plots
\usepackage{algorithm}
\usepackage{algpseudocode}
\usepackage{makecell}
\usepackage{bbding}
\usepackage{nopageno}

\usepackage[pagebackref=true,breaklinks=true,colorlinks,bookmarks=false]{hyperref}

\newcommand{\alg}[1]{Alg.~\ref{#1}}

\newcommand{\myparagraph}[1]{\vspace{-8pt} \paragraph{#1.}}

\def\cvprPaperID{4678} % *** Enter the CVPR Paper ID here
\def\confYear{CVPR 2021}

\begin{document}

%%%%%%%%% TITLE
\title{Bilevel Online Adaptation for Out-of-Domain Human Mesh Reconstruction}

\author{Shanyan Guan$^\ast$, 
Jingwei Xu\thanks{Equal contribution} , 
Yunbo Wang\thanks{Corresponding authors: Yunbo Wang, Bingbing Ni} , 
Bingbing Ni$^\dagger$, 
Xiaokang Yang \\
MoE Key Lab of Artificial Intelligence, AI Institute, Shanghai Jiao Tong University, China\\
{\tt\small \{shyanguan,xjwxjw,yunbow,nibingbing,xkyang\}@sjtu.edu.cn}
}
 
\maketitle

\begin{abstract}

This paper considers a new problem of adapting a pre-trained model of human mesh reconstruction to out-of-domain streaming videos. However, most previous methods based on the parametric SMPL model \cite{loper2015smpl} underperform in new domains with unexpected, domain-specific attributes, such as camera parameters, lengths of bones, backgrounds, and occlusions. Our general idea is to dynamically fine-tune the source model on test video streams with additional temporal constraints, such that it can mitigate the domain gaps without over-fitting the 2D information of individual test frames. A subsequent challenge is how to avoid conflicts between the 2D and temporal constraints. We propose to tackle this problem using a new training algorithm named Bilevel Online Adaptation (BOA), which divides the optimization process of overall multi-objective into two steps of weight probe and weight update in a training iteration. We demonstrate that BOA leads to state-of-the-art results on two human mesh reconstruction benchmarks\footnote{The project website with code, supplementary materials and video results is at \url{https://sites.google.com/view/humanmeshboa}}.

\end{abstract}
\section{Introduction}

\begin{figure}[t]
    \centering
    \includegraphics[width=\columnwidth]{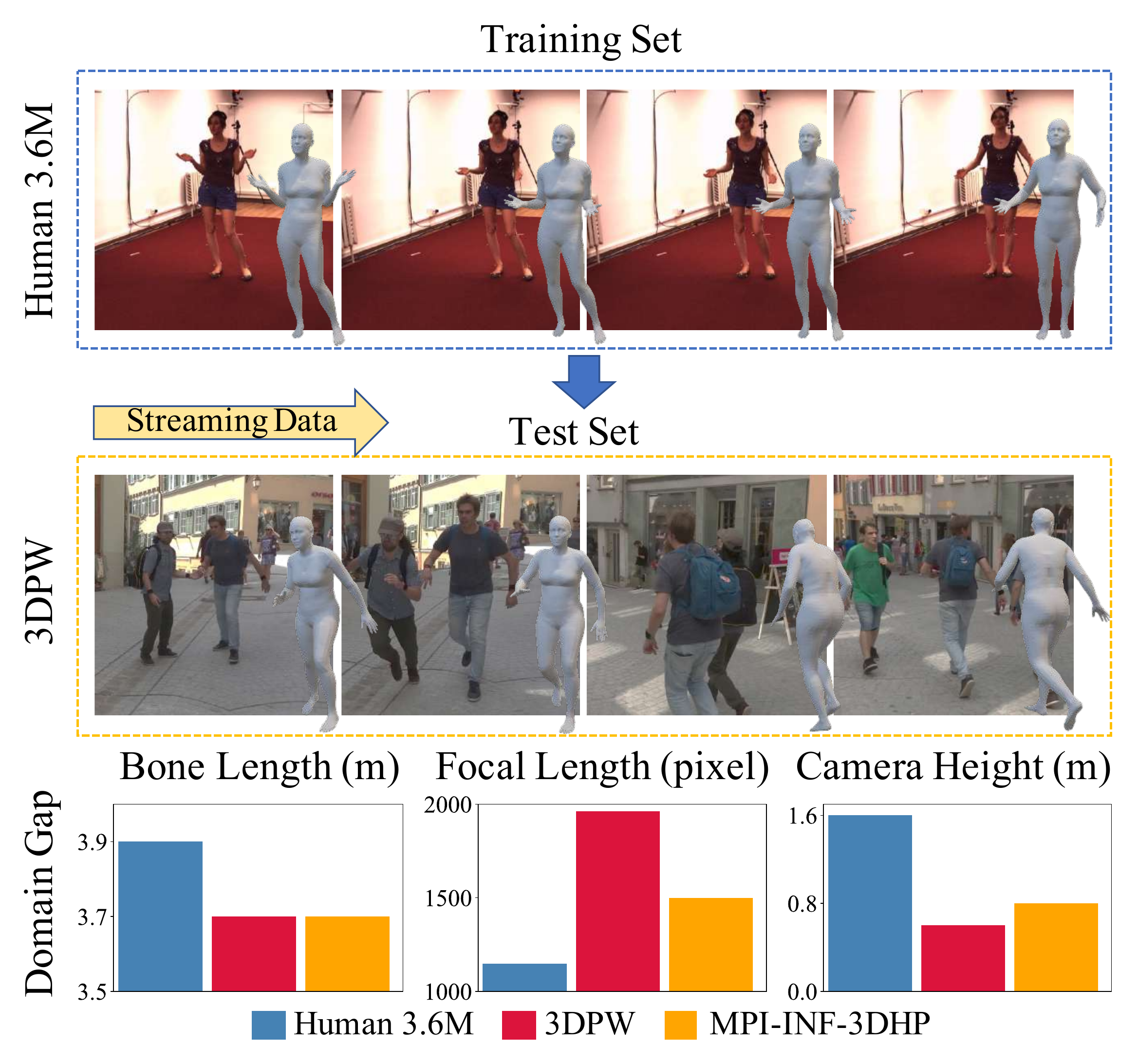}
    \caption{\textbf{Top:} Learning to reconstruct human meshes from out-of-domain streaming videos. The main challenges include the domain gaps of plain/crowded backgrounds, with/without occlusions, etc.
    \textbf{Bottom:} 
    Statistics of other typical domain gaps. Here, the bone length refers to the sum of lengths between body joints, whose topology is shared across datasets.}
    \label{fig:problem}
    \vspace{-10pt}
\end{figure}

Human mesh reconstruction is a hot topic in computer vision, where improving the generalization ability is one of the major challenges at present. We observe that previous models~\cite{kanazawa2018end,Moon_2020_ECCV_I2L-MeshNet,kocabas2020vibe,kolotouros2019learning,aksan2019structured,xu2019denserac,guler2019holopose,pavlakos2018learning} are prone to overfit the training dataset and usually underperform in out-of-domain testing scenarios.
As shown in Figure~\ref{fig:problem}, between different datasets, there usually exist large domain gaps in camera parameters, lengths of body bones, backgrounds, and occlusions, whose negative impact becomes even more severe when we apply the model to streaming data due to the rapidly changing environment of the test domain.
In this work, we are interested in finding a good solution to adapting human mesh reconstruction models to out-of-domain video frames that arrive in a sequential order, which is a practical task in many downstream, real applications, \eg, augmented reality~\cite{DBLP:conf/vr/Billinghurst04}, and human-robot interaction~\cite{DBLP:conf/hri/StubbsHW06}.

\begin{figure}[t]
    \centering
    \includegraphics[width=\linewidth]{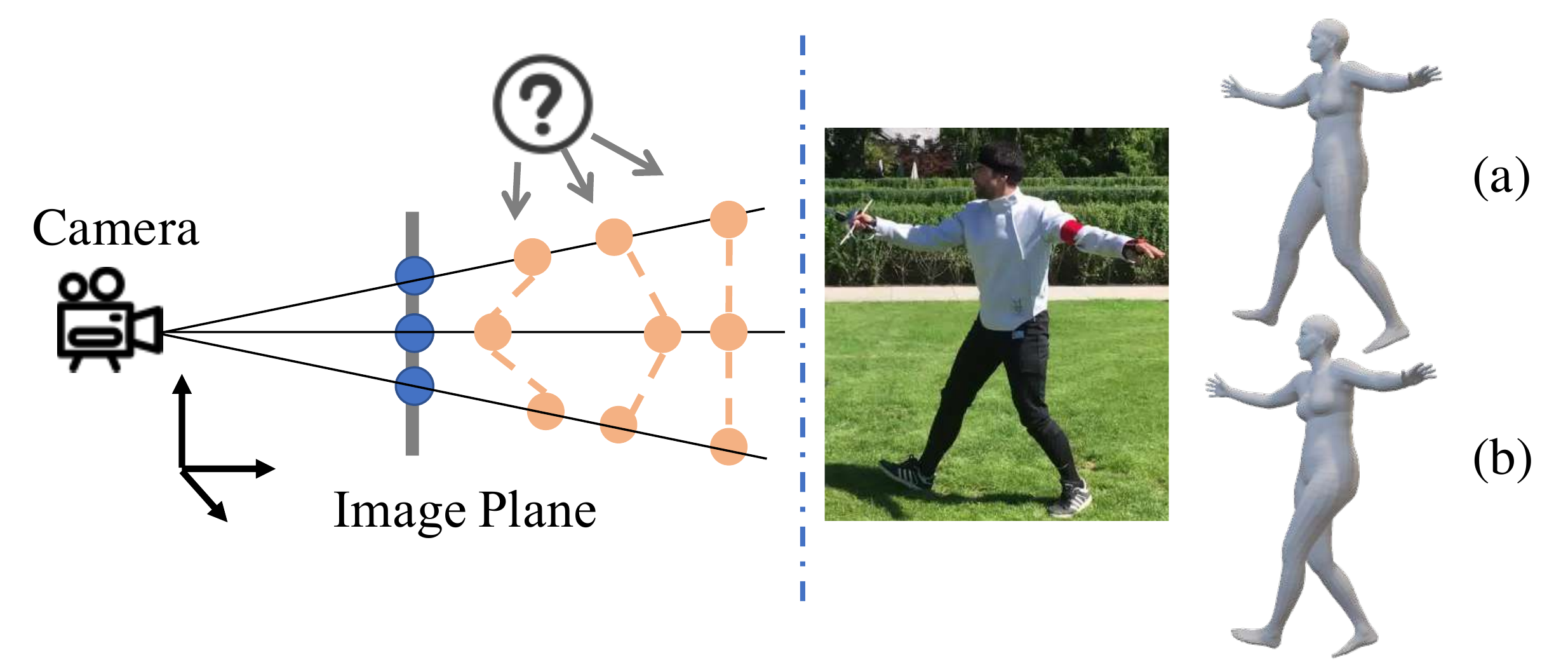}
    \caption{\textbf{Left:} Adapting the model with 2D re-projection loss leads to depth ambiguity, where multiple 3D vertices are projected to the same 2D position. \textbf{Right:} A showcase that the uncertainty of depth information may lead to the wrong estimation of leg posture.}
    \label{fig:motivation}
    \vspace{-5pt}
\end{figure}

The most serious technical challenge of this task is the lack of 3D annotations of test data. To cope with this problem, some optimization-based approaches~\cite{joo2020eft, loper2015smpl,SMPL-X:2019} learn to update the model on each test frame using frame-wise losses, such as the pose re-projection loss~\cite{kanazawa2018end,kolotouros2019learning} of 2D keypoints\footnote{It is a common practice to use the ground-truth 2D keypoints for cross-domain human mesh/pose reconstruction}. 
However, the imperfect frame-based loss functions do not always lead to effective online learning directions as expected by the 3D evaluation metric. There is a severe gap between them.  
As shown in Figure~\ref{fig:motivation}, it may cause severe ambiguity in the estimation of depth information, thus worsening the quality of mesh reconstruction.
Moreover, due to the asynchronous arrival of streaming data, the online adaptation model is prone to over-fitting, which will further amplify the difference between the 2D objectives and 3D evaluation metrics.

A straightforward solution is to regularize the training process towards 2D pose objectives using temporal constraints \cite{kocabas2020vibe,sun2019human,DBLP:conf/cvpr/KanazawaZFM19}, such as the smoothness of mesh reconstruction over time.
If the temporal constraints are used properly, the ambiguity of depth estimation can be greatly reduced.
However, empirically, a simple combination of 2D losses and temporal constraints tends to obtain undesirable results due to the competition and incompatibility between multiple objectives, in the sense that the gradient of 2D objectives may interfere with the training of the temporal one.
Further, solving this problem becomes even more urgent in online adaptation scenarios with streaming data, because without global knowledge of the test domain, the model can easily fall into a sub-optimal solution to either part of the loss functions that is more readily available.

The above two concerns motivate us to tackle the challenging problem of out-of-domain mesh reconstruction from a new perspective.
We propose an algorithm named \textit{Bilevel Online Adaptation} (BOA) that greatly benefits joint learning of multiple objectives in this task.
It effectively incorporates temporal consistency into the few-step online training by performing bilevel optimization on the streaming test data.
Specifically, in BOA, the lower-level optimization step serves as a weight probe to rational model parameters under single-frame pose constraints, while the upper-level optimization step finds a feasible response to overall loss function with temporal constraints. 
On one hand, our approach avoids overfitting the temporal constraints by retaining the 2D losses for the upper-level optimization. 
On the other hand, it avoids overfitting the 2D losses by updating the model only at the upper-level optimization step with second-order derivatives. 
By this means, our approach effectively combines the benefits of pose and temporal constraints.
In experiments, we use Human3.6M~\cite{h36m_pami} as the source domain, and take 3DPW~\cite{vonMarcard2018} and MPI-INF-3DHP~\cite{mehta2017monocular} as target domains with streaming video frames.
On both benchmarks, our approach consistently outperforms existing approaches~\cite{joo2020eft, loper2015smpl,SMPL-X:2019}, showing the excellent ability to tackle notable domain gaps. 
\section{Problem Setup}
\label{sec:problem}

%%%%%%%%%%%%%% Alg starts
\begin{algorithm*}[t] 
  \caption{Bilevel Online Adaptation}  
  \label{alg:Framwork}  
  \begin{algorithmic}[1]
    \Require  
    sequential frames $\{\boldsymbol{x}_i\}^{N}_{i=1}$ from test set, the base model $\mathcal{M}_{\phi_{0}}$ with model parameter $\phi_{0}$, a teacher model $\mathcal{T}_{\omega_0}$ with model parameter $\omega_0$, learning rates $\alpha$ and $\eta$.
    \Ensure  
    SMPL parameters $\{\widehat{\boldsymbol{\beta}}_{i}, \widehat{\boldsymbol{\theta}}_{i}\}^{N}_{i=1}$, camera parameters $\{\widehat{\boldsymbol{\psi}}_i\}^{N}_{i=1}$
    \State \textbf{Initialize} $\omega_{0} \leftarrow \phi_{0}$ 
    \For{$i=1, \ldots, N$}
        \State $\phi_{i-1}^{0} \leftarrow \phi_{i-1}$
        \For{$t=1, \ldots, T$}
        \Comment{For each bilevel optimization step}
            \State $\mathcal{L}_\text{low} \leftarrow \mathcal{L}_{F}(\boldsymbol{x}_{i};\phi_{i-1}^{t-1})$
            \Comment{Calculate lower-level loss with single-frame pose constraints}
            \State $\phi_{i-1}^{(t-1)\prime} \leftarrow \phi_{i-1}^{t-1} - \alpha \nabla_\phi \mathcal{L}_\text{low}$
            \Comment{Probe for rational weights under pose constraints}
            \State $\mathcal{L}_\text{up} \leftarrow \mathcal{L}_{F}(\boldsymbol{x}_i; \phi_{i-1}^{(t-1)\prime}) + \mathcal{L}_{T}(\boldsymbol{x}_i; \omega_{i-1}, \phi^{(t-1)\prime}_{i-1})$
            \Comment{Calculate upper-level loss with temporal constraints}
            \State $\phi^{t}_{i-1} \leftarrow \phi^{t-1}_{i-1} - \eta \nabla_\phi \mathcal{L}_\text{up}$
            \Comment{Update model weights with second-order derivatives}
        \EndFor
        \State $\phi_i \leftarrow \phi^{T}_{i-1}$
        \State $\omega_{i} \leftarrow \delta \omega_{i-1} + (1-\delta) \phi_{i}$
        \Comment{Update the teacher model}
        \State $\{\widehat{\boldsymbol{\beta}}_{i}, \widehat{\boldsymbol{\theta}}_{i}\}$, $\boldsymbol{\Pi}_{\widehat{\boldsymbol{\psi}}_i} \leftarrow \mathcal{M}_{\phi_{i}}(\boldsymbol{x}_{i})$
        \Comment{Estimate the SMPL parameters and the camera parameter of $\boldsymbol{x}_{i}$ using $\mathcal{M}_{\phi_i}$}
    \EndFor
  \end{algorithmic}  
\end{algorithm*}
%%%%%%%%%%%%%% Alg end

A SMPL-based solution to human mesh reconstruction can be usually specified as a tuple of $(\boldsymbol{X}, \boldsymbol{\Theta}, \boldsymbol{\Pi}, \mathcal{L})$, where $\boldsymbol{X}$ denotes the observation space\footnote{We here consider $\{\boldsymbol{x}_i\}_{i=1}^{N} \in \boldsymbol{X}$ as a set of consecutive video frames.}, and $\boldsymbol{\Theta}$ is the parameter space of SMPL \cite{loper2015smpl}. 
For each input frame $\boldsymbol{x}_i \in \boldsymbol{X}$, a first-stage model is trained to estimate $\{\widehat{\boldsymbol{\beta}}_i, \widehat{\boldsymbol{\theta}}_i\} \in \boldsymbol{\Theta}$. 
Then the SMPL model generates the corresponding mesh and recovers 3D keypoints denoted by $\widehat{\boldsymbol{J}}_i$ using a mesh-to-3D-skeleton mapping pre-defined in SMPL.
The third element in the tuple is a weak-perspective projection model for projecting $\widehat{\boldsymbol{J}}_i$ to 2D space, \ie, $\hat{\boldsymbol{j}}_i = \boldsymbol{\Pi}_{\widehat{\boldsymbol{\psi}}_i}(\widehat{\boldsymbol{J}}_i)$, where $\widehat{\boldsymbol{\psi}}_i$ is estimated from $\boldsymbol{x}_i$.
The last one in the tuple defines a loss function $\mathcal{L}(\cdot)$ on $(\widehat{\boldsymbol{\beta}}_i, \widehat{\boldsymbol{\theta}}_i, \widehat{\boldsymbol{\psi}}_i,\widehat{\boldsymbol{J}}_i, \widehat{\boldsymbol{j}}_i)$ to learn the first-stage model $\mathcal{M}_\phi$, usually in terms of neural networks.

In this work, we make two special modifications to the above task. First, we focus on out-of-domain scenarios, in the sense that large discrepancies may exist between the data distributions of the source training domain $\mathcal{D}^\text{tr}$ and the target test domain $\mathcal{D}^\text{test}$. 
Second, we specifically focus on dealing with streaming video frames at test time.
These changes bring in two challenges: 
(1) The ground truth values of the target domain parameters in $(\boldsymbol{\Theta}, \boldsymbol{\Pi})$ are always unavailable throughout the learning process, which is different from standard online learning.
(2) The distribution of the target domain is difficult to be estimated because the frames $\{\boldsymbol{x}_i\}_{i=1}^{N}\in\boldsymbol{X}$ are sequentially available and their distributions are continuously changing,
which is different from standard domain adaptation setups.
\section{Bilevel Online Adaptation}

In this section, we first formalize the online adaptation framework as a solution to out-of-domain human mesh reconstruction from streaming sequential data. 
To make the online adaptation more effective, we then propose a bilevel optimization algorithm that incorporates unsupervised temporal constraints into the training paradigm.

\subsection{Online Adaptation Framework}

Unlike the existing approaches~\cite{kanazawa2018end,DBLP:conf/nips/DoerschZ19,sun2019human} that try to solve the problem introduced in Section \ref{sec:problem} by learning more generalizable features in the source domain $\mathcal{D}^\text{tr}$, we here present an alternative solution that performs online test-time training directly on the target domain $\mathcal{D}^\text{test}$. A potential benefit is that as it is solely performed on $\mathcal{D}^\text{test}$, it can be jointly used with the state-of-the-art approaches that learn generalizable features from $\mathcal{D}^\text{tr}$ to further improve the quality of transfer learning.

\alg{alg:Framwork} shows the proposed online adaptation framework. Here we denote the pre-trained model from the source domain as $\mathcal{M}_{\phi_0}$. Our framework does not have special requirements for the pre-training method, but typically, $\mathcal{M}_{\phi_0}$ is trained offline to regress the ground truth SMPL parameters in a fully supervised manner. 
Given sequentially arrived target video frames $\{\boldsymbol{x}_i\}_{i=1}^N \in \mathcal{D}^\text{test}$, a straight forward solution to quickly absorbing the domain-specific knowledge is to fine-tuning $\mathcal{M}$ continuously on each individual $\boldsymbol{x}_i$, following the online adaptation paradigm proposed by Tonioni~\etal~\cite{tonioni2019real}. 
We take it as a baseline algorithm that computes the unsupervised loss function $\mathcal{L}$ with pose constraints\footnote{We discuss more about the specific forms of $\mathcal{L}$ in Section \ref{sec:expri}.} on each $\boldsymbol{x}_i$, and performs a single optimization step as follows before the inference step:
\begin{align}
    \phi_{i} = \phi_{i-1} - \alpha \nabla_\phi \mathcal{L}(\boldsymbol{x}_i; \mathcal{M}_{\phi_i}),
\end{align}
where $\alpha$ is the learning rate of gradient descent.

A potential disadvantage of the baseline algorithm is that although fine-tuning a learned model on unlabeled target data may help to handle rapidly changing test environments, an imperfect unsupervised loss function may lead to wrong directions of the one-step gradient descent, and may harm the overall algorithm. It may cause catastrophic overfitting to some undesirable information of current observation that is unrelated to the reconstruction quality.
To alleviate this issue, we propose the following spatiotemporal bilevel optimization approach.

\begin{figure*}[t]
    \centering
    \includegraphics[width=0.92\linewidth]{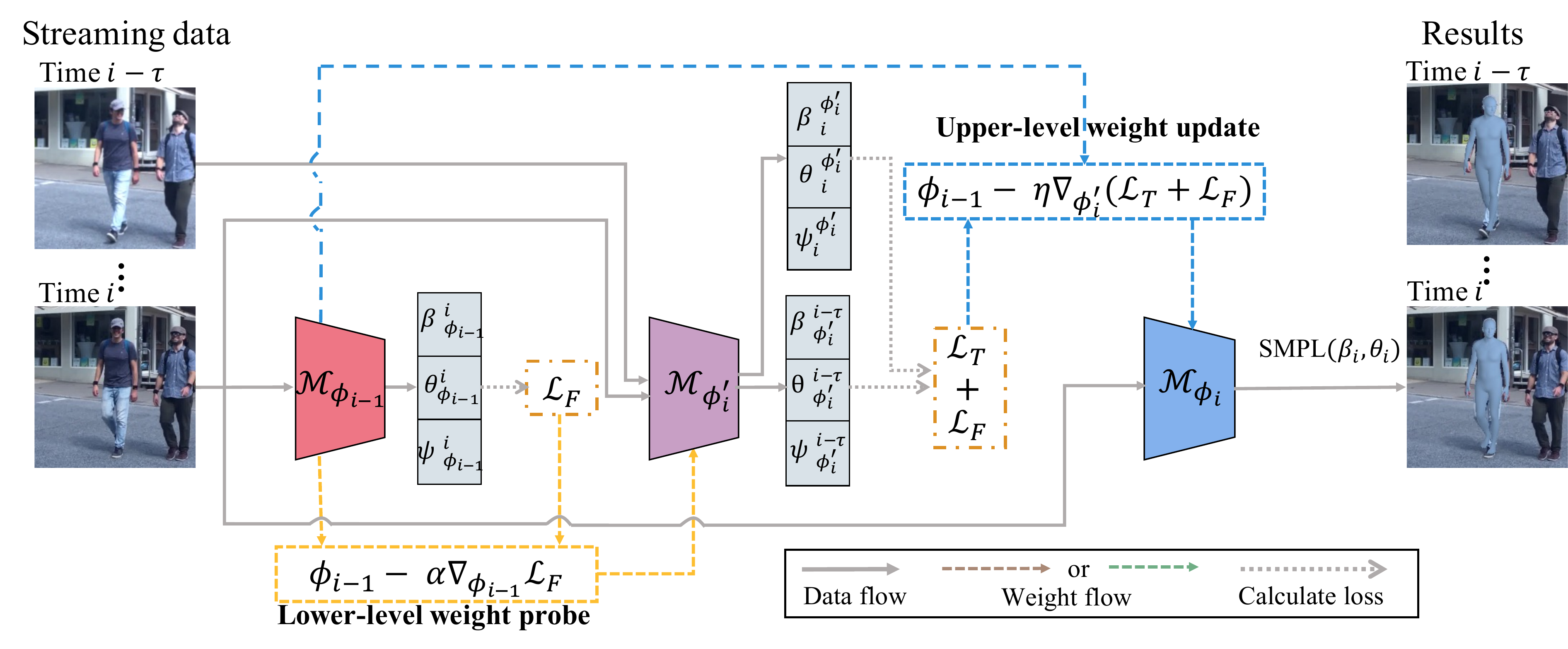}
    \caption{A diagram of the proposed bilevel online adaptation method. For simplicity, we only show one iteration of bilevel optimization. The lower-level training step serves as a parameter \textit{probe} to find a feasible response to the frame-wise pose constraints. The upper-level training step minimizes the overall multi-objectives in space-time and updates the model with second-order derivatives.
    }
    \vspace{-5pt}
    \label{fig:framework}
\end{figure*}

\subsection{Spatiotemporal Bilevel Optimization}

Considering the setting of out-of-domain streaming data, \ie, video frames arrive at a sequential order, there generally exists strong temporal dependency between frames, which can be leveraged to improve the quality of online adaptation.
Let us suppose that we have two objectives respectively for frame-wise constraints and temporal consistency, denoted by $\mathcal{L}_{F}$ and $\mathcal{L}_{T}$, whose specific forms will be discussed later. 
Straightforward approaches to combine $\mathcal{L}_{F}$ and $\mathcal{L}_{T}$ include jointly optimizing them by adding them together or iteratively performing two-stage optimization (Figure \ref{fig:motivation}\textcolor{red}{b}). 
However, these methods usually lead to sub-optimal results due to the competition and incompatibility between the objectives, in the sense that the gradient of the single-frame constraint may interfere with the training of the temporal one. 
We also observe that the single-frame constraint is usually optimized much faster than the temporal one. That is to say, in a small number of inference-stage optimization steps, the model may learn pose priors very quickly but then get stuck trying to learn temporal consistency.
Therefore, the first and foremost challenge we confront is to design an online optimization scheme to prevent overfitting to each objective and maximize the power of both single-frame and temporal constraints.

\myparagraph{Lower-level weight probe with single-frame constraints}
We formulate the problem of identifying effective model weights under spatiotemporal multi-objectives as a bilevel optimization problem. In this setup, as shown in Figure~\ref{fig:framework}, the lower-level optimization step serves as a weight probe to rational models under single-frame pose constraints, while the upper-level optimization step finds a feasible response to temporal constraints.
Specifically, for the $i$-th test sample, the model from the last online adaptation step, denoted by $\mathcal{M}_{\phi_{i-1}}$, is firstly optimized with the single-frame constraints, $\mathcal{L}_{F}$, to obtain a set of temporary weights denoted by $\phi_i^\prime$. 
We name this procedure as the \textit{lower-level probe} (\underline{Lines 5-6} in  \alg{alg:Framwork}), in the sense that first, $\phi_i^\prime$ can be feasible responses to the easy component of multi-objectives, which best facilitates the rest of the learning procedure for temporal consistency; Second, $\phi_i^\prime$ is not directly used to update $\mathcal{M}_{\phi_{i-1}}$.  
At this level we focus on the spatial constraints on individual frames:
\begin{align}
    \mathcal{L}_{F} = \gamma_1 ||\boldsymbol{j}_i - \widehat{\boldsymbol{j}_i}||_2^2 + \gamma_2 \rho(\widehat{\boldsymbol{\beta}}_i, \widehat{\boldsymbol{\theta}}_i) + \gamma_3 \mathcal{L}_{S},
    \label{Eq:streaming_adapt}
\end{align}
where $\{\gamma_1, \gamma_2, \gamma_3\}$ are the loss weights. The first term in $\mathcal{L}_{F}$ is a straightforward supervision of the re-projection error of 2D keypoints. The second term is the prior constraint on the shape and pose parameters, which is a common practice in mesh reconstruction. $\rho(\cdot)$ calculates the distance of the estimated $\widehat{\boldsymbol{\beta}}_i, \widehat{\boldsymbol{\theta}}_i$ to their statistic priors\footnote{These priors are obtained from a commonly-used third-party database.}. The third term is the fully supervised loss with 3D keypoints on a randomly sampled source data, which has two benefits: (1) preventing the catastrophic forgetting of the basic knowledge learned from $\mathcal{D}^\text{tr}$. (2) providing the online updated model a continuous 3D supervision to keep it from overfitting the imperfect unsupervised loss functions.  
After optimization at the lower level, we obtain the probe model $\mathcal{M}_{\phi_i^\prime}$ for subsequent upper-level learning. 
Note that, due to a lack of 3D supervisions in the target domain, the above $\mathcal{L}_{F}$ is insufficient to recover the 3D body. Therefore, it is essential to explore temporal correlations in streaming data to reduce the ambiguity of mesh construction.

\myparagraph{Upper-level weight update with temporal constraints}
At an upper optimization level, we calculate the overall spatiotemporal multi-objectives using $\mathcal{M}_{\phi_i^\prime}$ obtained at the lower-level optimization step, and then back-propagate with second-order derivatives to update the original $\phi_{i-1}$, as shown in \underline{Lines 7-8} in \alg{alg:Framwork}.
As for the specific form the motion constraints, given two images $\boldsymbol{x}_{i}$, $\boldsymbol{x}_{i-\tau}$ at an interval $\tau$ with their 2D keypoints $\boldsymbol{j}_i$, $\boldsymbol{j}_{i-\tau}$ and the estimated $\widehat{\boldsymbol{j}}_i$, $\widehat{\boldsymbol{j}}_{i-\tau}$, the motion loss is defined as
\begin{align}
    \mathcal{L}_{\boldsymbol{m}} &= ||\widehat{\boldsymbol{m}}_i - \boldsymbol{m}_i||_2^2,
\end{align}
where $\boldsymbol{m}_i = \boldsymbol{j}_i - \boldsymbol{j}_{i-\tau}$, and $\widehat{\boldsymbol{m}}_i = \widehat{\boldsymbol{j}}_i - \widehat{\boldsymbol{j}}_{i-\tau}$. Note that both $\widehat{\boldsymbol{j}}_i$ and $\widehat{\boldsymbol{j}}_{i-\tau}$ are obtained from the probe model $\mathcal{M}_{\phi_i^\prime}$.
Furthermore, we maintain an exponential moving average of history models with a teacher model~(similar to MeanTeacher~\cite{tarvainen2017mean}), denoted by $\mathcal{T}_\omega$. We regularize the output of $\mathcal{M}_{\phi_i^\prime}$ to be consistent with $\mathcal{T}_{\omega_{i-1}}$: 
\begin{align}
    \mathcal{L}_{\boldsymbol{mt}} = ||\mathcal{T}_{\omega_{i-1}}(\boldsymbol{x}) - \mathcal{M}_{\phi_i^\prime}(\boldsymbol{x})||_2^2,
\end{align}
which is then combined with the motion loss to obtain the overall temporal constraints that focus on the consistency of both the sequentially updated model weights and the reconstruction results as well:
\begin{align}
    \mathcal{L}_{T} = \mu_1 \mathcal{L}_{\boldsymbol{m}} + \mu_2 \mathcal{L}_{\boldsymbol{mt}},
\end{align}
where $\mu_1$ and $\mu_2$ control the weights of the two temporal loss terms. From another perspective, these two losses are complementary with each other: the teacher model $\mathcal{T}$ maintains long-term temporal information, and the motion loss $\mathcal{L}_{\boldsymbol{m}}$ is a constraint on short-term motion consistency.

\myparagraph{Alternatives for online adaptation schemes} 
As briefly mentioned above, there are several single-level optimization alternatives of the spatiotemporal multi-objectives, \eg, (1) one-stage joint adaptation: online adapting the model with a combined loss of $\mathcal{L}_{F}+\mathcal{L}_{T}$. (2) two-stage adaptation: adapting the model iteratively with $\mathcal{L}_{F}$ and $\mathcal{L}_{T}$ in a cascaded optimization manner.
However, we observe that the joint adaptation scheme is prone to lead to ineffective training of the temporal constraints due to the incompatibility between multiple objectives. The two-stage scheme adapts the model to individual frames under the single-frame constraints repeatedly, which commonly leads to severe overfitting and drifting away from the final 3D reconstruction metric.
The key insights of BOA are as follows: First, it avoids overfitting the temporal constraints by retaining the pose prior loss for the upper optimization level. 
Second, it avoids overfitting the pose priors by updating the model weights only at the upper optimization level with second-order derivatives. 
By this means, BOA effectively combines the profits of both single-frame and temporal constraints, achieving considerable improvement over its alternatives.

\myparagraph{Network architectures}
Following the majority of previous SMPL-based human reconstruction models, we use a ResNet-50~\cite{he2016deep} pre-trained on ImageNet~\cite{deng2009imagenet} for encoding individual video frames. The encoded features are then delivered to two fully-connected layers with $1{,}024$ neurons, followed by a dropout layer~\cite{srivastava2014dropout}. The final layer of $\mathcal{M}_\phi$ is a fully-connected layer with $85$ neurons. During streaming adaptation, only one image is taken as input. As a result, we replace Batch Normalization~\cite{ioffe2015batch} with Group Normalization~\cite{wu2018group} to estimate more accurate statistics.

\section{Experiments}
\label{sec:expri}

\paragraph{Datasets.} 
We use the Human3.6M dataset for training the source model and learn to adapt the model to the 3DHP and 3DPW datasets. 
Table~\ref{tab:dataset_gap} presents the statistics of typical domain gaps among these datasets.

\begin{itemize}
    \item \textbf{Human3.6M \cite{h36m_pami}} is captured in a controlled environment, which has $11$ subjects in total. Following the previous approaches~\cite{kocabas2020vibe,kanazawa2018end}, we train the base model on $5$ subjects (S1, S5, S6, S7, S8), and down-sample all videos from $50$fps to $10$fps.
    \item \textbf{3DHP \cite{mehta2017monocular}} is the test split of the MPI-INF-3DHP dataset. It consists of $2{,}929$ valid frames from $6$ subjects performing $7$ actions, collected from both indoor and outdoor environments.
    \item \textbf{3DPW \cite{vonMarcard2018}} is a multi-person dataset captured by a handheld camera, where most videos are collected from outdoor environments. As 3DHP, we also use the test set of 3DPW as a streaming target domain.
\end{itemize}

\begin{table}[t]
    \centering
    \resizebox{\linewidth}{!}{
    \footnotesize
    \begin{tabular}{lcccc}
    \toprule
         Dataset      &\makecell[c]{Focal \\ len. (pixel)}       &\makecell[c]{Bone \\ len. (m)}    &\makecell[c]{Camera \\dist. (m)}   &\makecell[c]{Camera\\ ht. (m)}  \\
    \midrule
         H3.6M   &1146.8$\pm$2.0   &3.9$\pm$0.1  &5.2$\pm$0.8     &1.6$\pm$0.1   \\
         3DPW         &1962.2$\pm$1.5   &3.7$\pm$0.1  &3.5$\pm$0.7     &0.6$\pm$0.8    \\
         3DHP &1497.9$\pm$2.8   &3.7$\pm$0.1  &3.8$\pm$0.8     &0.8$\pm$0.4    \\
    \bottomrule
    \end{tabular}
    }
    \vspace{2pt}
    \caption{Typical domain gaps among datasets in terms of focal length, bone length, camera distance, and camera height~\cite{DBLP:journals/corr/abs-2004-03143}.}
    \vspace{-5pt}
    \label{tab:dataset_gap}
\end{table}

\begin{figure*}[t]
    \centering
    \includegraphics[width=0.9\textwidth]{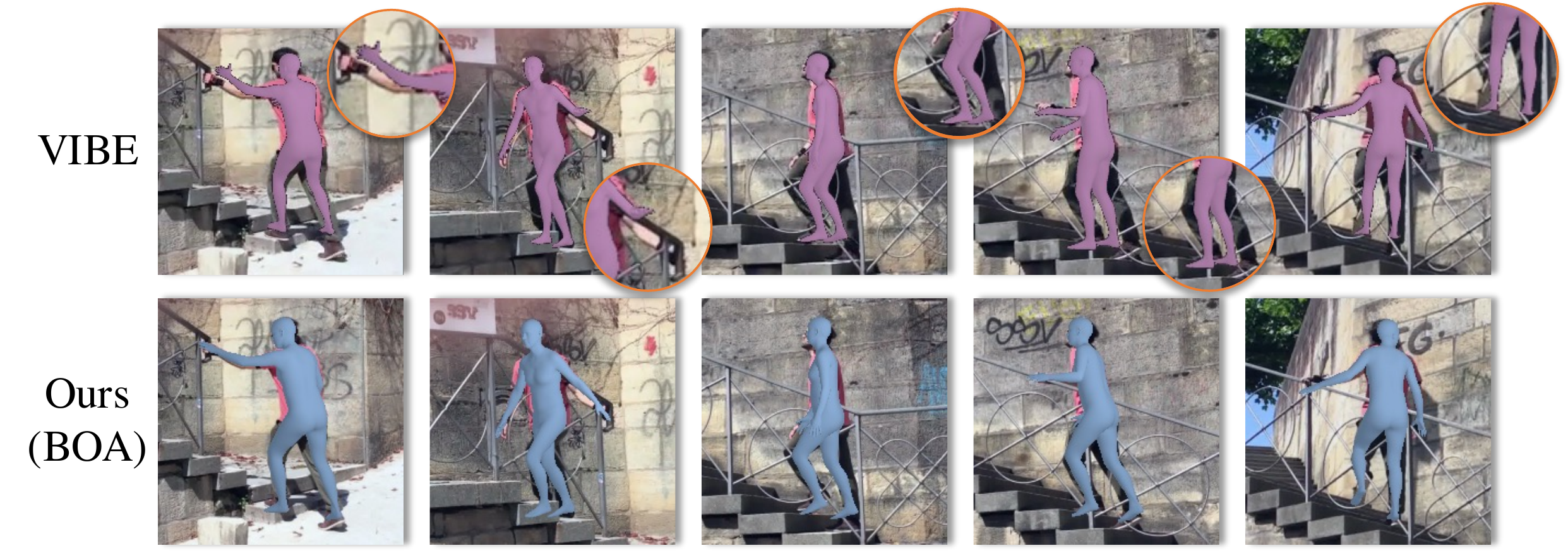}
    \vspace{-3pt}
        \caption{A qualitative comparison of mesh reconstruction on 3DPW streaming data. We zoom in on the limbs for better visualization.}
    \label{fig:comparison}
    \vspace{-8pt}
\end{figure*}

\myparagraph{Training details} We first train the base model $\mathcal{M}$ on the Human3.6M dataset and take 3DPW and 3DHP as test sets. All video frames are cropped and then scaled to $224\times224$ pixels according to the bounding boxes calculated from 2D keypoints. 
For the base model $\mathcal{M}$, we follow the same training scheme as SPIN (more details can be found in \cite{kolotouros2019learning}).
For the training of BOA on 3DPW, we choose the Adam optimizer~\cite{kingma2014adam} with the learning rate $\eta=3e^{-6}$ ($\beta_1=0.5, \beta_2=0.999$). The loss weights in $\mathcal{L}_{F}$ are $\gamma_1=10, \gamma_2=1$, and $\gamma_3=0.1$. The loss weights in $\mathcal{L}_{T}$ are $\mu_1=0.1$ and $\mu_2=0.1$. 
As for 3DHP, the learning rate is set to $2e^{-6}$ ($\beta_1=0.2, \beta_2=0.999$). We set $\gamma_1=10, \gamma_2=1, \gamma_3=0.1$ in $\mathcal{L}_{F}$ and  $\mu_1=0.1, \mu_2=0.1$ in $\mathcal{L}_{T}$.
Note that the order of streaming videos in 3DPW and 3DHP is pre-defined (same for all compared models), and the batch size of online optimization is $1$.
We set $T=1$ (\alg{alg:Framwork}) for the efficiency of adaptation.
Please refer to the supplementary material for more analyses of hyper-parameters.

\myparagraph{Baselines} We initially compare BOA with end-to-end methods, including frame-based methods \cite{kanazawa2018end,kolotouros2019convolutional,Choi_2020_ECCV_Pose2Mesh,kolotouros2019learning,Choi_2020_ECCV_Pose2Mesh,Moon_2020_ECCV_I2L-MeshNet}, video-based methods~\cite{DBLP:conf/cvpr/KanazawaZFM19,kocabas2020vibe}, and those attempting to learn generalizable features from the training domain, such as Sim2Real~\cite{DBLP:conf/nips/DoerschZ19} and DSD-SATN~\cite{sun2019human}. Given a video frame, end-to-end methods directly estimate its SMPL parameters.
We also include existing approaches that fine-tune SMPL parameters $\boldsymbol{\beta, \theta}$~\cite{bogo2016keep,arnab2019exploiting} or model parameters $\phi$~\cite{joo2020eft} on the target domain.
Different from these approaches, BOA adapts $\phi_{i}$ in an online fashion, which is more challenging. Please refer to the supplementary material for more details.

\myparagraph{Evaluation metrics} Following previous works~\cite{kanazawa2018end,DBLP:conf/nips/DoerschZ19,zhang2020inference}, we evaluate our model in terms of Mean Per Joint Position Error (MPJPE), Procrustes-Aligned MPJPE (PA-MPJPE), and the Percentage of Correct Keypoints (PCK) with a threshold of $150$mm on 3DHP.

\begin{table}[t]
\centering
\small
    \begin{tabular}{lccc}
    \toprule 
        Method & Prot.    &PA-MPJPE$^\downarrow$ &MPJPE$^\downarrow$    \\
    \midrule
        HMR~\cite{kanazawa2018end}       & \#PH      &76.7     &130.0  \\
        Sim2Real~\cite{DBLP:conf/nips/DoerschZ19}   & \#PH      &74.7     &-      \\
        GraphCMR~\cite{kolotouros2019convolutional}    & \#PS      &70.2     &-      \\
        SPIN~\cite{kolotouros2019learning}        & \#PS      &59.2     &96.9   \\
        I2L-MeshNet~\cite{Moon_2020_ECCV_I2L-MeshNet} & \#PS      &58.6     &93.2   \\
        Pose2Mesh~\cite{Choi_2020_ECCV_Pose2Mesh}   & \#PS      &58.9     &89.2   \\
        DSD-SATN~\cite{sun2019human}    & \#PS      &69.5     &-      \\    % training set is only h36m
        HMMR~\cite{DBLP:conf/cvpr/KanazawaZFM19}        & \#PH      &73.6     &116.5  \\

    \midrule
        SMPLify~\cite{bogo2016keep}                                & \#PH      &106.1    &199.2  \\
        Arnab~\etal~\cite{arnab2019exploiting} & \#PH      &72.2     &-      \\
        EFT~\cite{joo2020eft}         & \#PS      &55.7     &-      \\
    \midrule
        BOA          & \#PH      &58.8     &92.1  \\
        BOA         & \#PS      &\textbf{49.5}    &\textbf{77.2}      \\
\bottomrule
\end{tabular}
% }
\vspace{5pt}
\caption{Results on 3DPW, including end-to-end approaches (top) and those fine-tuned on the target domain (middle).}
\label{tab:3dpw}
\vspace{-5pt}
\end{table}

\begin{table}[t]
    \centering
    \small
    \begin{tabular}{lcc}
    \toprule
     Prot. & SMPL annotation & \#Valid frames \\
\midrule
     \#PS (SPIN)   & Original            & 35,515   \\
     \#PH (HMMR)   & The fits            & 26,234   \\
    \bottomrule
    \end{tabular}
    \vspace{5pt}
    \caption{Different protocols of pre-processing the 3DPW data by SPIN \cite{kolotouros2019learning} and HMMR \cite{DBLP:conf/cvpr/KanazawaZFM19}. \#PS uses the SMPL annotations from the original 3DPW as labels, while \#PH uses the fitted neutral results (without gender information) as labels.}
    \label{tab:setting_diff}
    \vspace{-5pt}
\end{table}

\subsection{Quantitative Evaluation}

\paragraph{Results on 3DPW.} Table~\ref{tab:3dpw} presents quantitative comparisons on 3DPW in MPJPE and PA-MPJPE. Following HMMR or SPIN, most existing methods adopt two kinds of pre-processing protocols on 3DPW as illustrated in Table~\ref{tab:setting_diff}. 
These two protocols have significant differences in the number of test images and SMPL annotations, which have a great impact on the evaluation. Please refer to the supplementary materials for more details. 
In Table~\ref{tab:3dpw}, we mark the protocol used in the original literature of each compared method. 
Compared with other end-to-end methods (top part), BOA achieves better performance in both \#PS and \#PH, and particularly outperforms the methods that are designed to learn generalizable features at training time~\cite{sun2019human, kanazawa2018end, DBLP:conf/nips/DoerschZ19}, which indicates that our test-time adaptation approach can better mitigate the domain gap by properly exploiting the streaming data from the test domain. 
Besides, we also observe that BOA outperforms the compared models that are fine-tuned on the entire training set of 3DPW in an offline manner (middle part). Note that BOA does not require access to the training set.
In addition, we do not include the results from VIBE \cite{kocabas2020vibe} ($56.5$ in PA-MPJPE and $93.5$ in MPJPE) in quantitative comparison, since it was evaluated on the same number of test images under \#PH but uses the same SMPL annotations under \#PS.

\begin{table}[t]
    \centering
    \small
    \begin{tabular}{lcccc}
    \toprule
         Method              & 3DHP-$\mathcal{D}^\text{train}$           &PA-MPJPE$^\downarrow$    &MPJPE$^\downarrow$ &PCK$^\uparrow$  \\
    \midrule
         Vnect            &\Checkmark                              &98.0        &124.7  &83.9 \\
         HMR  &\Checkmark                              &113.2       &169.5  &77.1 \\
         SPIN &\Checkmark                              &80.4        &124.8  &87.0 \\
         BOA                &\XSolidBrush                            &\textbf{77.4}            &\textbf{117.6}       &\textbf{90.3}     \\
    \bottomrule
    \end{tabular}
    \vspace{5pt}
    \caption{Results on 3DHP. All models but BOA are trained on the training split of MPI-INF-3DHP, while BOA performs best.}
    \label{tab:3dhp}
    \vspace{-5pt}
\end{table}

\myparagraph{Results on 3DHP} 
Table~\ref{tab:3dhp} gives the MPJPE and PA-PMJPE results on 3DHP, which is the test set of the entire MPI-INF-3DHP domain. 
Note that all models but BOA are directly trained on the training set $\mathcal{D}^{\text{train}}$ of MPI-INF-3DHP in an offline fashion, in the sense that the global knowledge from the test domain is more accessible to these compared models. 
Although BOA has never been trained on $\mathcal{D}^{\text{train}}$, it still performs best on the corresponding test split, showing a strong adaptability to a rapidly changing test environment.

\begin{figure*}[t]
    \centering
    \includegraphics[width=0.9\linewidth]{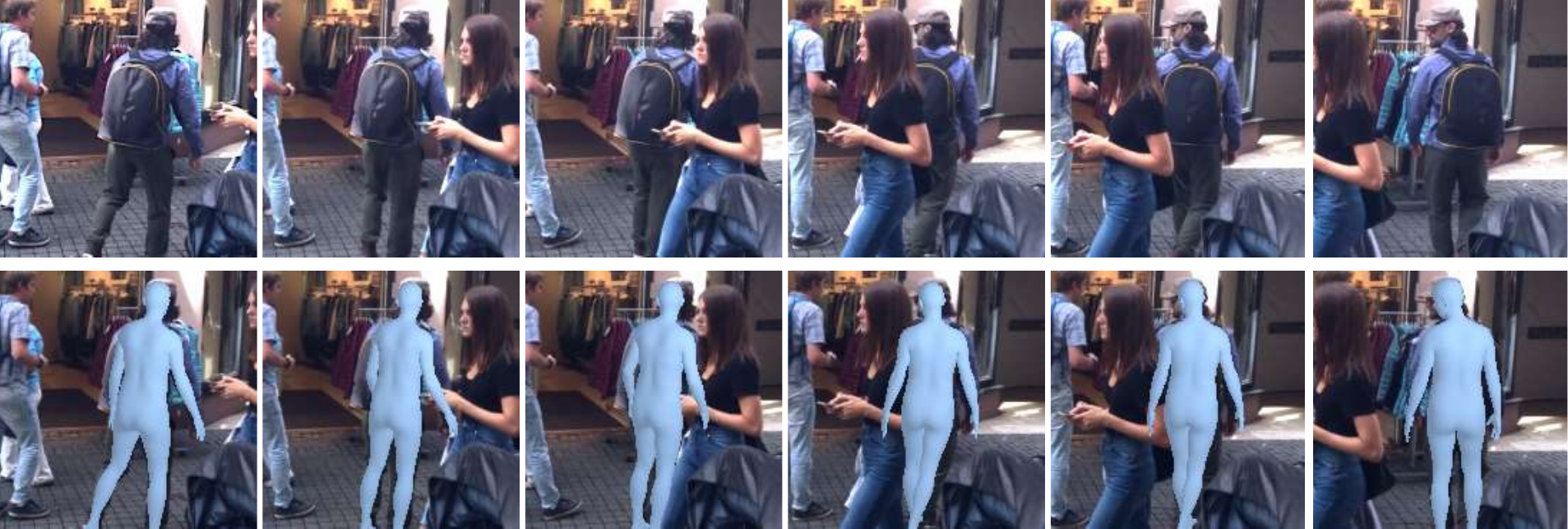}
    \caption{Reconstruction results under severe occlusion. First row: the input frames. Second row: mesh reconstruction results. The subject of interest (the man in the middle) is occluded by the walking woman, which is challenging for out-of-domain mesh reconstruction.}
    \label{fig:occlusion_analysis}
    \vspace{-5pt}
\end{figure*}

\subsection{Qualitative Evaluation}

Figure~\ref{fig:comparison} presents a typical showcase of mesh reconstruction on the challenging 3DPW dataset.
The first row refers to human meshes generated by VIBE~\cite{kocabas2020vibe}, while the second row corresponds to our results. 
We zoom in on the limbs for better visualization and observe that the reconstruction quality of VIBE is less satisfying, \eg, the positions of arms and legs are not correctly estimated.
By contrast, our model can capture the depth structure of the human subject, which is mainly due to the proposed bilevel optimization scheme and spatiotemporal constraints.
Figure~\ref{fig:occlusion_analysis} presents a sequence of input videos with severe occlusions, where the subject of interest (the man in the middle) is covered by the walking woman. Still, BOA successfully estimates the occluded human body.

\begin{figure}[t]
    \centering
    \includegraphics[width=\linewidth]{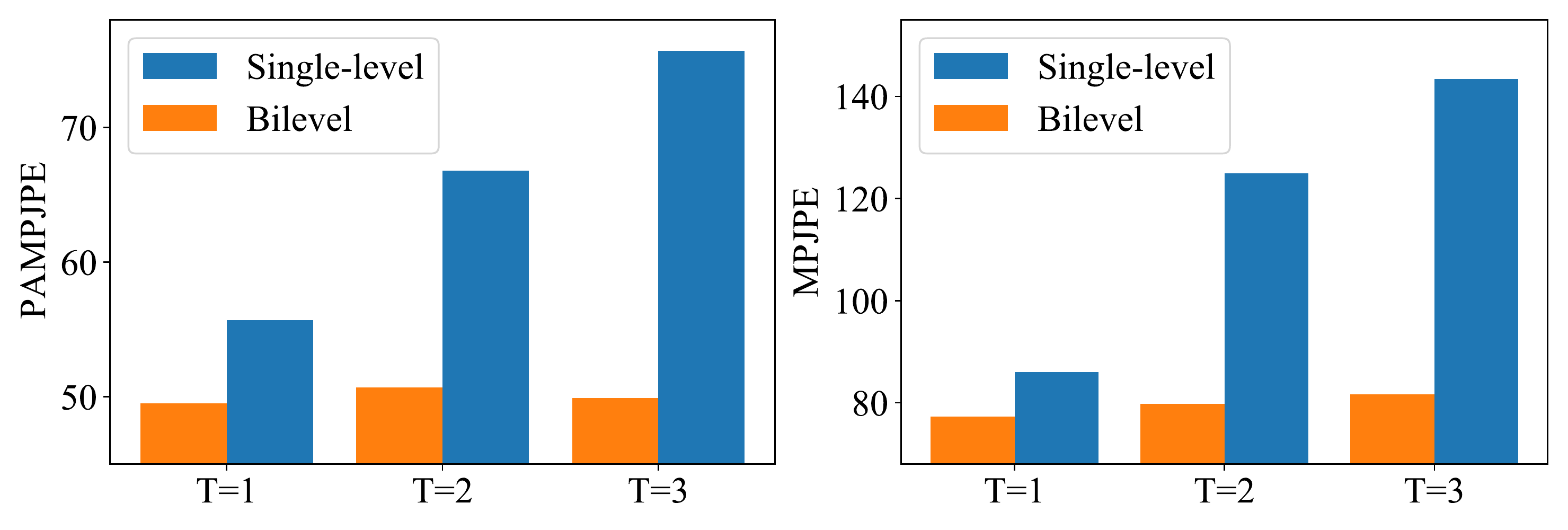}
    \caption{Ablations on the number of optimization steps $T$ (\#PS on 3DPW). BOA has a more stable performance with the growth of $T$, compared with the its single-level counterpart.}
    \label{fig:overfit_ablates}
    \vspace{-10pt}
\end{figure}

\subsection{Ablation Study}
\label{sec:abla}

\paragraph{Ablations on the number of optimization steps.}
With the growth of the optimization steps ($T$ in \alg{alg:Framwork}), as shown in Figure~\ref{fig:overfit_ablates}, the error of single-level training scheme, which combines $\mathcal{L}_F$ and $\mathcal{L}_T$ in a multi-objective, in both PA-MPJPE and MPJPE increases quickly. %
For comparison, the performance of BOA decreases at a much slower rate, which indicates that the single-level optimization is more likely to result in over-fitting to the current video frame, and thus makes it difficult for the model to quickly adapt to the next frame. This effect can be greatly alleviated by the proposed bilevel optimization method.

\begin{table}[t]
    \centering
    % \footnotesize
    \resizebox{\linewidth}{!}{
    \begin{tabular}{llllcc}
    \toprule
    Index               &Optim.           &$\mathcal{L}_\text{low}$                    &$\mathcal{L}_\text{up}$    &PA-MPJPE &MPJPE  \\
    \midrule
    \textbf{B1}       &1-stage        &$\mathcal{L}_{F}$                    &-                    &55.7 &86.0   \\
    \textbf{B2}       &1-stage        &$\mathcal{L}_{T}$                    &-                    &140.1 &245.5   \\      
    \textbf{B3}       &1-stage        &$\mathcal{L}_{F},\mathcal{L}_{T}$    &-                    &58.9 &94.5   \\ 
    \midrule
    \textbf{B4}       &2-stage        &$\mathcal{L}_{F}$                    &$\mathcal{L}_{T}$    &59.3   &91.5   \\   
    \textbf{B5}       &2-stage        &$\mathcal{L}_{F}$                    &$\mathcal{L}_{F},\mathcal{L}_{T}$   &55.2 &85.1   \\  
    \midrule
    \textbf{B6}       &Bilevel                       &$\mathcal{L}_{F}$                    &$\mathcal{L}_{T}$    &142.5 &257.1    \\   
    \textbf{Final}    &Bilevel                       &$\mathcal{L}_{F}$                    &$\mathcal{L}_{F},\mathcal{L}_{T}$   &49.5 &77.3    \\   
    \bottomrule
    \end{tabular}
    }
    \vspace{2pt}
    \caption{Ablation studies on the training schemes (\#PS on 3DPW). \textbf{B1-B3} update model parameters with constant forms of losses in a single optimization step. \textbf{B4-B5} update model parameters with alternate loss functions. \textbf{B6} and \textbf{Final} use bilevel optimization.}
    \label{tab:bilevel_ablates}
\end{table}

\myparagraph{Analyses on the online bilevel optimization framework} 
As shown in Table~\ref{tab:bilevel_ablates}, we investigate the effectiveness of the proposed bilevel adaptation scheme and compare it with other variants.
Specifically, \textbf{B1-B3} refer to the single-level, one-stage optimization scheme, while \textbf{B4-B5} are trained by updating model parameters with alternate loss functions (\ie, $\mathcal{L}_\text{low}$ and $\mathcal{L}_\text{up}$).
Note that the major difference between two-stage and bilevel is whether the parameters in $\mathcal{M}_{\phi_{i}}$ are obtained from $\mathcal{M}_{\phi_{i-1}^{\prime}}$ or $\mathcal{M}_{\phi_{i-1}}$.
We observe that, despite the use of temporal constraints, \textbf{B3} performs worse than \textbf{B1}, indicating that the straightforward combination of multi-objectives leads to sub-optimal results.
Even though we use the multi-objectives in a two-stage training scheme (\textbf{B5}), we can only observe a minor improvement over the vanilla \textbf{B1} model.
By contrast, the final proposed BOA is shown to effectively combine the best of both constraints and achieve considerable improvement over all compared baselines.

\begin{table}[t]
    \centering
    \small
    \begin{tabular}{lccccc}
    \toprule
         Index              &$\mathcal{L}_{F}$      & $\mathcal{L}_{\boldsymbol{m}}$     & $\mathcal{L}_{\boldsymbol{mt}}$   &PA-MPJPE         &MPJPE\\
    \midrule
         \textbf{B7}      &\Checkmark             &\XSolidBrush           &\Checkmark        &53.0    &81.8                    \\
         \textbf{B8}      &\Checkmark             &\Checkmark             &\XSolidBrush      &51.7     &82.1                   \\
         \textbf{Final}   &\Checkmark             &\Checkmark             &\Checkmark        &49.5    &77.3                   \\
    \bottomrule 
    \end{tabular}
    \vspace{5pt}
    \caption{Ablations on BOA temporal constraints (\#PS on 3DPW).}
    \vspace{-5pt}
    \label{tab:temporal_ablates}
\end{table}

\myparagraph{Ablations on temporal constraints}
Table~\ref{tab:temporal_ablates} shows the ablation studies for the proposed two temporal constraints $\mathcal{L}_{\boldsymbol{mt}}$ and $\mathcal{L}_{\boldsymbol{m}}$. 
By comparing \textbf{B7} with \textbf{Final}, we can observe that the ues of $\mathcal{L}_{\boldsymbol{m}}$ reduces PA-MPJPE from $53.0$mm to $49.5$mm, while the use of $\mathcal{L_{T}}$ \textbf{B8} reduces PA-MPJPE from $51.7$mm to $49.5$mm. 
A possible reason is that the motion loss $\mathcal{L}_{\boldsymbol{m}}$ focuses on short-term temporal constraint and helps to recalibrate pose artifacts relative to the last frame. 
By comparing \textbf{B8} with \textbf{Final}, we can find that the use of $\mathcal{L}_{\boldsymbol{mt}}$ significantly reduces MPJPE, which indicates that the long-term information carried by $\mathcal{L}_{\boldsymbol{mt}}$ is beneficial for consistent mesh reconstruction. It may help to mitigate the domain gaps caused by systematic biases such as the focal length and camera orientations.

\begin{figure}[t]
    \centering
    \includegraphics[width=\linewidth]{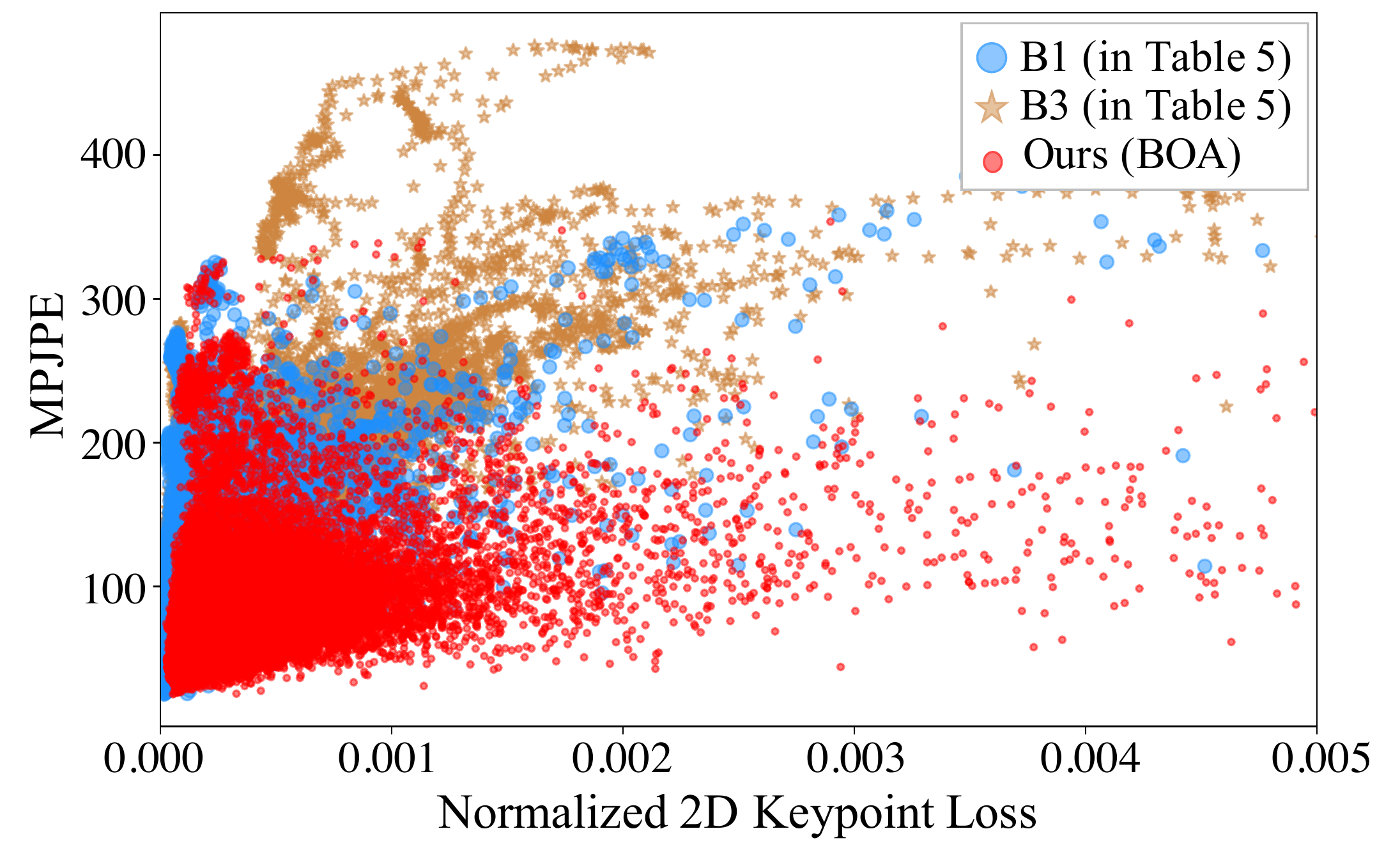}
    \vspace{-15pt}
    \caption{Correlations between the 2D pose re-projection loss and the evaluation metric. 
    \textbf{B1} uses frame-based losses only. 
    \textbf{B3} optimizes the frame-based and temporal losses by integrating them in a unified loss function.
    Only with \textbf{BOA}, the MPJPE results are consistent with the frame-based loss, showing that BOA greatly reduces the 3D ambiguity of estimated meshes.}
    \label{fig:loss-metric}
    \vspace{-5pt}
\end{figure}

\myparagraph{Ablations on loss-metric correlations}
In Figure~\ref{fig:loss-metric}, the X-axis refers to the normalized 2D keypoint loss and the Y-axis is the MPJPE. 
The blue dots are the results of the vanilla baseline only trained with frame-based loss functions (\textbf{B1}), the yellow stars correspond to the baseline model trained with multi-objectives (\textbf{B3}), and the red dots indicate bilevel online adaptation. 
We can see that, first, although a straightforward use of temporal constraints (\textbf{B3}) achieves comparable results in the 2D loss with \textbf{B1}, it harms the 3D evaluation metric.
Second, with BOA, the model achieves MPJPE results that are more consistent with the frame-based loss, indicating that BOA can reduce 3D ambiguity by using the temporal constraints more appropriately.

\section{Related Work}
\paragraph{3D human mesh reconstruction.}
SMPL~\cite{loper2015smpl} is a widely used parametric model for 3D human mesh reconstruction, which is also adopted in this work.
The early methods~\cite{guan2009estimating,sigal2008combined,bogo2016keep,lassner2017unite,huang2017towards,zanfir2018monocular} generally adopt the optimization scheme, where a standard T-Pose SMPL model is gradually fit to an input image according to the silhouettes~\cite{lassner2017unite} or 2D keypoints~\cite{bogo2016keep}.
These optimization-based methods are time-consuming, \ie, they often struggle to reduce the inference time spent on a single input.
Recently, many approaches~\cite{kanazawa2018end,Moon_2020_ECCV_I2L-MeshNet,kocabas2020vibe,kolotouros2019learning,aksan2019structured,xu2019denserac,guler2019holopose,pavlakos2018learning} use deep neural networks to regress the parameters of the SMPL model, which are efficient and accurate if large-scale data is available.
The major drawback of CNN-based regression models is the generalization ability.
For example, deep models trained on indoor dataset generally do not have satisfying results~\cite{kanazawa2018end} if tested on an in-the-wild dataset.
To tackle this problem, Kanazawa~\etal~\cite{kanazawa2018end} propose an adversarial framework, utilizing the unpaired 3D annotations, to facilitate the reconstruction. 
Several researches~\cite{sun2019human,tung2017self,li2019towards,omran2018neural,rueegg2020chained} also show that the paired 3D annotation is not necessary, attempting to find more representative temporal features~\cite{sun2019human,tung2017self} or employ more informative input such as RGB-D~\cite{li2019towards}, and part segmentation~\cite{omran2018neural,rueegg2020chained} to facilitate human mesh reconstruction.
However, there still exists a principled challenge in this task, where neither the unpaired 3D annotation nor the other mentioned intermediate representations could effectively fill the gap between two largely different datasets.
In this work, we propose to tackle this problem by using an online adaptation algorithm, named BOA. The key insight is BOA exploits the time constraints of test frames while avoiding overfitting with bilevel optimization.

\myparagraph{Unsupervised online adaptation}
Unsupervised online adaptation refers to sequentially adapting a pre-trained model at test time in an unsupervised manner. It is an emerging technique to prevent model crashing when the test data is diverse from the training data.
Previous methods~\cite{duchi2011adaptive,broderick2013streaming,bobu2018adapting,liu2020learning,park2018meta,voigtlaender2017online,tonioni2019real,broderick2013streaming,tonioni2019learning,zhang2020online,li2020self} use it for tasks other than mesh reconstruction, such as video segmentation~\cite{voigtlaender2017online}, tracking~\cite{park2018meta}, and stereo matching~\cite{tonioni2019real,tonioni2019learning}. 
In this paper, we present a pilot study of unsupervised online adaptation in the context of human mesh reconstruction. 
Beyond unsupervised online adaptation, many previous approaches effectively learn generalizable features through meta-learning \cite{finn2017model,fallah2020convergence}, extracting domain-invariant representations \cite{khosla2012undoing,muandet2013domain,ghifary2015domain,li2017deeper,li2018domain,li2018learning,li2019episodic}, or learning with adversarial examples \cite{shankar2018generalizing,volpi2018generalizing} without requiring access to target labels. %
However, none of these approaches focus on how to adapt a pre-trained model to streaming data.

\section{Conclusion}

In this paper, we presented a new research problem of reconstructing human meshes from out-of-domain streaming videos. 
We proposed a new online adaption algorithm that learns temporal consistency with bilevel optimization and demonstrated that it can greatly benefit the multi-objective training process in space-time.
Our approach outperforms the state-of-the-art mesh reconstruction methods on two benchmarks with rapidly changing test environments. 

\section*{Acknowledgement}
This work was supported by NSFC (U19B2035) and Shanghai Municipal Science \& Technology Major Project (2021SHZDZX0102). This work was also supported by the NSFC grants U20B2072 and 61976137. 

{\small
\bibliographystyle{ieee_fullname}
\bibliography{egbib}

\begin{thebibliography}{10}\itemsep=-1pt

\bibitem{aksan2019structured}
Emre Aksan, Manuel Kaufmann, and Otmar Hilliges.
\newblock Structured prediction helps 3{D} human motion modelling.
\newblock In {\em ICCV}, pages 7144--7153, 2019.

\bibitem{arnab2019exploiting}
Anurag Arnab, Carl Doersch, and Andrew Zisserman.
\newblock Exploiting temporal context for 3d human pose estimation in the wild.
\newblock In {\em CVPR}, pages 3395--3404, 2019.

\bibitem{DBLP:conf/vr/Billinghurst04}
Mark Billinghurst.
\newblock Introduction to {A}ugmented {R}eality.
\newblock In {\em VR}, page 266, 2004.

\bibitem{bobu2018adapting}
Andreea Bobu, Eric Tzeng, Judy Hoffman, and Trevor Darrell.
\newblock Adapting to continuously shifting domains.
\newblock In {\em ICLR}, 2018.

\bibitem{bogo2016keep}
Federica Bogo, Angjoo Kanazawa, Christoph Lassner, Peter Gehler, Javier Romero,
  and Michael~J Black.
\newblock Keep it {SMPL}: Automatic estimation of 3d human pose and shape from
  a single image.
\newblock In {\em ECCV}, pages 561--578, 2016.

\bibitem{broderick2013streaming}
Tamara Broderick, Nicholas Boyd, Andre Wibisono, Ashia~C Wilson, and Michael~I
  Jordan.
\newblock Streaming variational bayes.
\newblock In {\em NeurIPS}, pages 1727--1735, 2013.

\bibitem{Choi_2020_ECCV_Pose2Mesh}
Hongsuk Choi, Gyeongsik Moon, and Kyoung~Mu Lee.
\newblock Pose2{M}esh: Graph convolutional network for 3d human pose and mesh
  recovery from a 2d human pose.
\newblock In {\em ECCV}, pages 769--787, 2020.

\bibitem{deng2009imagenet}
Jia Deng, Wei Dong, Richard Socher, Li-Jia Li, Kai Li, and Li Fei-Fei.
\newblock Image{N}et: A large-scale hierarchical image database.
\newblock In {\em CVPR}, pages 248--255, 2009.

\bibitem{DBLP:conf/nips/DoerschZ19}
Carl Doersch and Andrew Zisserman.
\newblock Sim2real transfer learning for 3d human pose estimation: motion to
  the rescue.
\newblock In {\em NeurIPS}, pages 12949--12961, 2019.

\bibitem{duchi2011adaptive}
John Duchi, Elad Hazan, and Yoram Singer.
\newblock Adaptive subgradient methods for online learning and stochastic
  optimization.
\newblock {\em Journal of machine learning research}, 2011.

\bibitem{fallah2020convergence}
Alireza Fallah, Aryan Mokhtari, and Asuman Ozdaglar.
\newblock On the convergence theory of gradient-based model-agnostic
  meta-learning algorithms.
\newblock In {\em AISTATS}, pages 1082--1092, 2020.

\bibitem{finn2017model}
Chelsea Finn, Pieter Abbeel, and Sergey Levine.
\newblock Model-agnostic meta-learning for fast adaptation of deep networks.
\newblock In {\em ICML}, pages 1126--1135, 2017.

\bibitem{ghifary2015domain}
Muhammad Ghifary, W Bastiaan~Kleijn, Mengjie Zhang, and David Balduzzi.
\newblock Domain generalization for object recognition with multi-task
  autoencoderss.
\newblock In {\em ICCV}, pages 2551--2559, 2015.

\bibitem{guan2009estimating}
Peng Guan, Alexander Weiss, Alexandru~O Balan, and Michael~J Black.
\newblock Estimating human shape and pose from a single image.
\newblock In {\em ICCV}, pages 1381--1388, 2009.

\bibitem{guler2019holopose}
Riza~Alp Guler and Iasonas Kokkinos.
\newblock Holopose: {H}olistic 3{D} human reconstruction in-the-wild.
\newblock In {\em CVPR}, pages 10884--10894, 2019.

\bibitem{he2016deep}
Kaiming He, Xiangyu Zhang, Shaoqing Ren, and Jian Sun.
\newblock Deep residual learning for image recognition.
\newblock In {\em CVPR}, pages 770--778, 2016.

\bibitem{huang2017towards}
Yinghao Huang, Federica Bogo, Christoph Lassner, Angjoo Kanazawa, Peter~V
  Gehler, Javier Romero, Ijaz Akhter, and Michael~J Black.
\newblock Towards accurate marker-less human shape and pose estimation over
  time.
\newblock In {\em 3DV}, pages 421--430, 2017.

\bibitem{ioffe2015batch}
Sergey Ioffe and Christian Szegedy.
\newblock Batch {N}ormalization: Accelerating deep network training by reducing
  internal covariate shift.
\newblock In {\em ICML}, pages 448--456, 2015.

\bibitem{h36m_pami}
Catalin Ionescu, Dragos Papava, Vlad Olaru, and Cristian Sminchisescu.
\newblock Human3.6{M}: Large scale datasets and predictive methods for 3d human
  sensing in natural environments.
\newblock {\em IEEE transactions on pattern analysis and machine intelligence},
  pages 1325--1339, 2014.

\bibitem{joo2020eft}
Hanbyul Joo, Natalia Neverova, and Andrea Vedaldi.
\newblock Exemplar {F}ine-{T}uning for 3d human pose fitting towards
  in-the-wild 3d human pose estimation.
\newblock {\em arXiv preprint arXiv:2004.03686}, 2020.

\bibitem{kanazawa2018end}
Angjoo Kanazawa, Michael~J Black, David~W Jacobs, and Jitendra Malik.
\newblock End-to-end recovery of human shape and pose.
\newblock In {\em CVPR}, pages 7122--7131, 2018.

\bibitem{DBLP:conf/cvpr/KanazawaZFM19}
Angjoo Kanazawa, Jason~Y. Zhang, Panna Felsen, and Jitendra Malik.
\newblock Learning 3d human dynamics from video.
\newblock In {\em CVPR}, pages 5614--5623, 2019.

\bibitem{khosla2012undoing}
Aditya Khosla, Tinghui Zhou, Tomasz Malisiewicz, Alexei~A Efros, and Antonio
  Torralba.
\newblock Undoing the damage of dataset bias.
\newblock In {\em ECCV}, pages 158--171, 2012.

\bibitem{kingma2014adam}
Diederik Kingma and Jimmy Ba.
\newblock Adam: A method for stochastic optimization.
\newblock In {\em ICLR}, 2015.

\bibitem{kocabas2020vibe}
Muhammed Kocabas, Nikos Athanasiou, and Michael~J Black.
\newblock {VIBE}: Video inference for human body pose and shape estimation.
\newblock In {\em CVPR}, pages 5253--5263, 2020.

\bibitem{kolotouros2019learning}
Nikos Kolotouros, Georgios Pavlakos, Michael~J Black, and Kostas Daniilidis.
\newblock Learning to reconstruct 3d human pose and shape via model-fitting in
  the loop.
\newblock In {\em ICCV}, pages 2252--2261, 2019.

\bibitem{kolotouros2019convolutional}
Nikos Kolotouros, Georgios Pavlakos, and Kostas Daniilidis.
\newblock Convolutional mesh regression for single-image human shape
  reconstruction.
\newblock In {\em CVPR}, pages 4501--4510, 2019.

\bibitem{lassner2017unite}
Christoph Lassner, Javier Romero, Martin Kiefel, Federica Bogo, Michael~J
  Black, and Peter~V Gehler.
\newblock Unite the people: Closing the loop between 3{D} and 2{D} human
  representations.
\newblock In {\em CVPR}, pages 6050--6059, 2017.

\bibitem{li2017deeper}
Da Li, Yongxin Yang, Yi-Zhe Song, and Timothy~M Hospedales.
\newblock Deeper, broader and artier domain generalization.
\newblock In {\em ICCV}, pages 5542--5550, 2017.

\bibitem{li2018learning}
Da Li, Yongxin Yang, Yi-Zhe Song, and Timothy~M Hospedales.
\newblock Learning to generalize: Meta-learning for domain generalization.
\newblock In {\em AAAI}, 2018.

\bibitem{li2019episodic}
Da Li, Jianshu Zhang, Yongxin Yang, Cong Liu, Yi-Zhe Song, and Timothy~M
  Hospedales.
\newblock Episodic training for domain generalization.
\newblock In {\em ICCV}, pages 1446--1455, 2019.

\bibitem{li2018domain}
Haoliang Li, Sinno Jialin~Pan, Shiqi Wang, and Alex~C Kot.
\newblock Domain generalization with adversarial feature learning.
\newblock In {\em CVPR}, pages 5400--5409, 2018.

\bibitem{li2019towards}
Ren Li, Changjiang Cai, Georgios Georgakis, Srikrishna Karanam, Terrence Chen,
  and Ziyan Wu.
\newblock Towards robust {RGB-D} human mesh recovery.
\newblock {\em arXiv preprint arXiv:1911.07383}, 2019.

\bibitem{li2020self}
Shunkai Li, Xin Wang, Yingdian Cao, Fei Xue, Zike Yan, and Hongbin Zha.
\newblock Self-supervised deep visual odometry with online adaptation.
\newblock In {\em CVPR}, pages 6339--6348, 2020.

\bibitem{liu2020learning}
Hong Liu, Mingsheng Long, Jianmin Wang, and Yu Wang.
\newblock Learning to {A}dapt to {E}volving {D}omains.
\newblock In {\em NeurIPS}, 2020.

\bibitem{loper2015smpl}
Matthew Loper, Naureen Mahmood, Javier Romero, Gerard Pons-Moll, and Michael~J
  Black.
\newblock {SMPL}: A skinned multi-person linear model.
\newblock {\em ACM Transactions on Graphics (TOG)}, pages 1--16, 2015.

\bibitem{mehta2017monocular}
Dushyant Mehta, Helge Rhodin, Dan Casas, Pascal Fua, Oleksandr Sotnychenko,
  Weipeng Xu, and Christian Theobalt.
\newblock Monocular 3{D} human pose estimation in the wild using improved cnn
  supervision.
\newblock In {\em 3DV}, pages 506--516, 2017.

\bibitem{Moon_2020_ECCV_I2L-MeshNet}
Gyeongsik Moon and Kyoung~Mu Lee.
\newblock I2l-{M}esh{N}et: Image-to-lixel prediction network for accurate 3d
  human pose and mesh estimation from a single rgb image.
\newblock In {\em ECCV}, 2020.

\bibitem{muandet2013domain}
Krikamol Muandet, David Balduzzi, and Bernhard Sch{\"o}lkopf.
\newblock Domain generalization via invariant feature representation.
\newblock In {\em ICML}, pages 10--18, 2013.

\bibitem{omran2018neural}
Mohamed Omran, Christoph Lassner, Gerard Pons-Moll, Peter Gehler, and Bernt
  Schiele.
\newblock Neural body fitting: Unifying deep learning and model based human
  pose and shape estimation.
\newblock In {\em 3DV}, pages 484--494, 2018.

\bibitem{park2018meta}
Eunbyung Park and Alexander~C Berg.
\newblock Meta-tracker: Fast and robust online adaptation for visual object
  trackers.
\newblock In {\em ECCV}, pages 569--585, 2018.

\bibitem{SMPL-X:2019}
Georgios Pavlakos, Vasileios Choutas, Nima Ghorbani, Timo Bolkart, Ahmed A.~A.
  Osman, Dimitrios Tzionas, and Michael~J. Black.
\newblock Expressive {B}ody {C}apture: 3{D} {H}ands, {F}ace, and {B}ody from a
  {S}ingle {I}mage.
\newblock In {\em CVPR}, pages 10975--10985, 2019.

\bibitem{pavlakos2018learning}
Georgios Pavlakos, Luyang Zhu, Xiaowei Zhou, and Kostas Daniilidis.
\newblock Learning to estimate 3d human pose and shape from a single color
  image.
\newblock In {\em CVPR}, pages 459--468, 2018.

\bibitem{rueegg2020chained}
Nadine Rueegg, Christoph Lassner, Michael~J Black, and Konrad Schindler.
\newblock Chained representation cycling: Learning to estimate 3d human pose
  and shape by cycling between representations.
\newblock In {\em AAAI}, pages 5561--5569, 2020.

\bibitem{shankar2018generalizing}
Shiv Shankar, Vihari Piratla, Soumen Chakrabarti, Siddhartha Chaudhuri, Preethi
  Jyothi, and Sunita Sarawagi.
\newblock Generalizing across domains via cross-gradient training.
\newblock In {\em ICLR}, 2018.

\bibitem{sigal2008combined}
Leonid Sigal, Alexandru Balan, and Michael~J Black.
\newblock Combined discriminative and generative articulated pose and non-rigid
  shape estimation.
\newblock In {\em NeurIPS}, pages 1337--1344, 2008.

\bibitem{srivastava2014dropout}
Nitish Srivastava, Geoffrey Hinton, Alex Krizhevsky, Ilya Sutskever, and Ruslan
  Salakhutdinov.
\newblock Dropout: a simple way to prevent neural networks from overfitting.
\newblock {\em The journal of machine learning research}, pages 1929--1958,
  2014.

\bibitem{DBLP:conf/hri/StubbsHW06}
Kristen Stubbs, Pamela~J. Hinds, and David Wettergreen.
\newblock Challenges to grounding in human-robot interaction.
\newblock In {\em HRI}, pages 357--358, 2006.

\bibitem{sun2019human}
Yu Sun, Yun Ye, Wu Liu, Wenpeng Gao, YiLi Fu, and Tao Mei.
\newblock Human mesh recovery from monocular images via a skeleton-disentangled
  representation.
\newblock In {\em ICCV}, pages 5349--5358, 2019.

\bibitem{tarvainen2017mean}
Antti Tarvainen and Harri Valpola.
\newblock Mean teachers are better role models: Weight-averaged consistency
  targets improve semi-supervised deep learning results.
\newblock In {\em NeurIPS}, pages 1195--1204, 2017.

\bibitem{tonioni2019learning}
Alessio Tonioni, Oscar Rahnama, Thomas Joy, Luigi~Di Stefano, Thalaiyasingam
  Ajanthan, and Philip~HS Torr.
\newblock Learning to adapt for stereo.
\newblock In {\em CVPR}, pages 9661--9670, 2019.

\bibitem{tonioni2019real}
Alessio Tonioni, Fabio Tosi, Matteo Poggi, Stefano Mattoccia, and Luigi~Di
  Stefano.
\newblock Real-time self-adaptive deep stereo.
\newblock In {\em CVPR}, pages 195--204, 2019.

\bibitem{tung2017self}
Hsiao-Yu Tung, Hsiao-Wei Tung, Ersin Yumer, and Katerina Fragkiadaki.
\newblock Self-supervised learning of motion capture.
\newblock In {\em NeurIPS}, pages 5242--5252, 2017.

\bibitem{voigtlaender2017online}
Paul Voigtlaender and Bastian Leibe.
\newblock Online adaptation of convolutional neural networks for video object
  segmentation.
\newblock In {\em BMVC}, 2017.

\bibitem{volpi2018generalizing}
Riccardo Volpi, Hongseok Namkoong, Ozan Sener, John~C Duchi, Vittorio Murino,
  and Silvio Savarese.
\newblock Generalizing to unseen domains via adversarial data augmentation.
\newblock In {\em NeurIPS}, 2018.

\bibitem{vonMarcard2018}
Timo von Marcard, Roberto Henschel, Michael Black, Bodo Rosenhahn, and Gerard
  Pons-Moll.
\newblock Recovering accurate 3d human pose in the wild using {IMU}s and a
  moving camera.
\newblock In {\em ECCV}, pages 601--617, 2018.

\bibitem{DBLP:journals/corr/abs-2004-03143}
Zhe Wang, Daeyun Shin, and Charless~C. Fowlkes.
\newblock Predicting camera viewpoint improves cross-dataset generalization for
  3d human pose estimation.
\newblock In {\em ECCV}, pages 523--540, 2020.

\bibitem{wu2018group}
Yuxin Wu and Kaiming He.
\newblock Group {N}ormalization.
\newblock In {\em ECCV}, pages 3--19, 2018.

\bibitem{xu2019denserac}
Yuanlu Xu, Song-Chun Zhu, and Tony Tung.
\newblock Denserac: Joint 3{D} pose and shape estimation by dense
  render-and-compare.
\newblock In {\em ICCV}, pages 7760--7770, 2019.

\bibitem{zanfir2018monocular}
Andrei Zanfir, Elisabeta Marinoiu, and Cristian Sminchisescu.
\newblock Monocular 3{D} pose and shape estimation of multiple people in
  natural scenes-the importance of multiple scene constraints.
\newblock In {\em CVPR}, pages 2148--2157, 2018.

\bibitem{zhang2020inference}
Jianfeng Zhang, Xuecheng Nie, and Jiashi Feng.
\newblock Inference stage optimization for cross-scenario 3{D} human pose
  estimation.
\newblock In {\em NeurIPS}, 2020.

\bibitem{zhang2020online}
Zhenyu Zhang, Stephane Lathuiliere, Elisa Ricci, Nicu Sebe, Yan Yan, and Jian
  Yang.
\newblock Online depth learning against forgetting in monocular videos.
\newblock In {\em CVPR}, pages 4494--4503, 2020.

\end{thebibliography}
}

\end{document}